\documentclass[letterpaper, 10 pt, conference]{ieeeconf}
\usepackage{times}
\usepackage{graphicx}
\usepackage{amsmath,amssymb,amsopn,amstext,amsfonts}
\usepackage{cancel}
\usepackage[space]{cite}
\usepackage{pdfsync}
\usepackage{balance}
\usepackage{color}
\usepackage{mathtools}
\usepackage{algpseudocode}
\usepackage{algorithm} 
\usepackage{bm}

\usepackage{diagbox}
\usepackage{float}
\usepackage{epstopdf}
\usepackage{pifont}
\usepackage{fixltx2e}
\usepackage{amsmath}
\usepackage{multirow}
\usepackage{url}
\usepackage{verbatim}
\makeatletter
\let\NAT@parse\undefined
\makeatother
\usepackage[linkcolor=black,citecolor=black,urlcolor=black,colorlinks=TRUE]{hyperref}
\usepackage{booktabs}
\usepackage{graphicx}
\usepackage{subcaption}
\usepackage{threeparttable}  
\usepackage{gensymb}
\graphicspath{{./figures/}}
\DeclareGraphicsExtensions{.png,.jpg,.eps,.pdf}
\IEEEoverridecommandlockouts
\begin{document}
	\title{\LARGE \bf Model-Based Planning and Control for Terrestrial-Aerial Bimodal Vehicles with Passive Wheels}
	\author{Ruibin Zhang, Junxiao Lin, Yuze Wu, Yuman Gao, Chi Wang, Chao Xu, Yanjun Cao, and Fei Gao
	\thanks{This work was supported by the Fundamental Research Funds for the Central Universities and the National Natural Science Foundation of China under grant no.62003299. All authors are with the State Key Laboratory of Industrial Control Technology, Zhejiang University, Hangzhou 310027, China, and also with the Huzhou Institute of Zhejiang University, Huzhou 313000, China.} 
	\thanks{Corresponding author: Fei Gao.}
	\thanks{E-mail:{\tt\small \{ruibin\_zhang, fgaoaa\}@zju.edu.cn}}}

	\maketitle
	\thispagestyle{empty}
	\pagestyle{empty}
	\begin{abstract}
	\label{sec:abstract}\textbf{}Terrestrial and aerial bimodal vehicles have gained widespread attention due to their cross-domain maneuverability. Nevertheless, their bimodal dynamics significantly increase the complexity of motion planning and control, thus hindering robust and efficient autonomous navigation in unknown environments. To resolve this issue, we develop a model-based planning and control framework for terrestrial aerial bimodal vehicles. This work begins by deriving a unified dynamic model and the corresponding differential flatness. Leveraging differential flatness, an optimization-based trajectory planner is proposed, which takes into account both solution quality and computational efficiency. Moreover, we design a tracking controller using nonlinear model predictive control based on the proposed unified dynamic model to achieve accurate trajectory tracking and smooth mode transition. We validate our framework through extensive benchmark comparisons and experiments, demonstrating its effectiveness in terms of planning quality and control performance.
	
	\end{abstract} 
	
	\IEEEpeerreviewmaketitle

	\section{Introduction}
    \label{sec:Introduction}

    Mobile robots play a prominent role in many aspects of human society in recent years. Among them, unmanned aerial vehicles are widely used in unstructured and cluttered environments due to their high maneuverability, but they are not suitable for long-distance missions due to poor energy efficiency. In contrast, while unmanned ground vehicles enjoy much higher energy efficiency, their ability to maneuver over terrains with unavoidable obstacles is 
    limited. In order to combine the advantages of both types of vehicles, researchers have devoted considerable effort to developing terrestrial-aerial bimodal vehicles (TABVs), which feature cross-domain locomotion\cite{kalantari2013design, dudley2015micro, morton2017small, mintchev2018multi, sabet2019rollocopter, li2021driving, choi2021baxter, david2021design, kim2021bipedal, jia2022quadrotor}. Moreover, some previous works present autonomous navigation frameworks for TABVs, trying to make them competent for autonomous tasks in complex scenarios.
    
   TABVs have different dynamic characteristics in different locomotion modes, which pose significant challenges to trajectory planning and motion control. First, regarding trajectory planning, it is vital to ensure dynamical feasibility. However, due to the TABVs' bimodal nature, it is difficult to find a proper formulation to guarantee the dynamic feasibility of both modes, as well as the continuity and smoothness during mode transition. In addition, it is also necessary to improve planning efficiency for real-time performance. Previous works \cite{araki2017multi, sharif2018energy, choudhury2019dynamic, fan2019autonomous, zhang2022autonomous, wu2023unified} either oversimplify the dynamics of TABV or suffer from high computational consumption. In terms of motion control, previous efforts\cite{colmenares2019nonlinear, fan2019autonomous, zhang2022autonomous, yang2022sytab} mainly aim at developing a unified control framework for both modes. Nonetheless, the nonlinear dynamics of both modes are partially ignored for the convenience of modeling, thus restricting control performance in aggressive scenarios where the linearization or approximation substantially limits the solution space.

	 \begin{figure}
	\centering
	\includegraphics[width=1\linewidth]{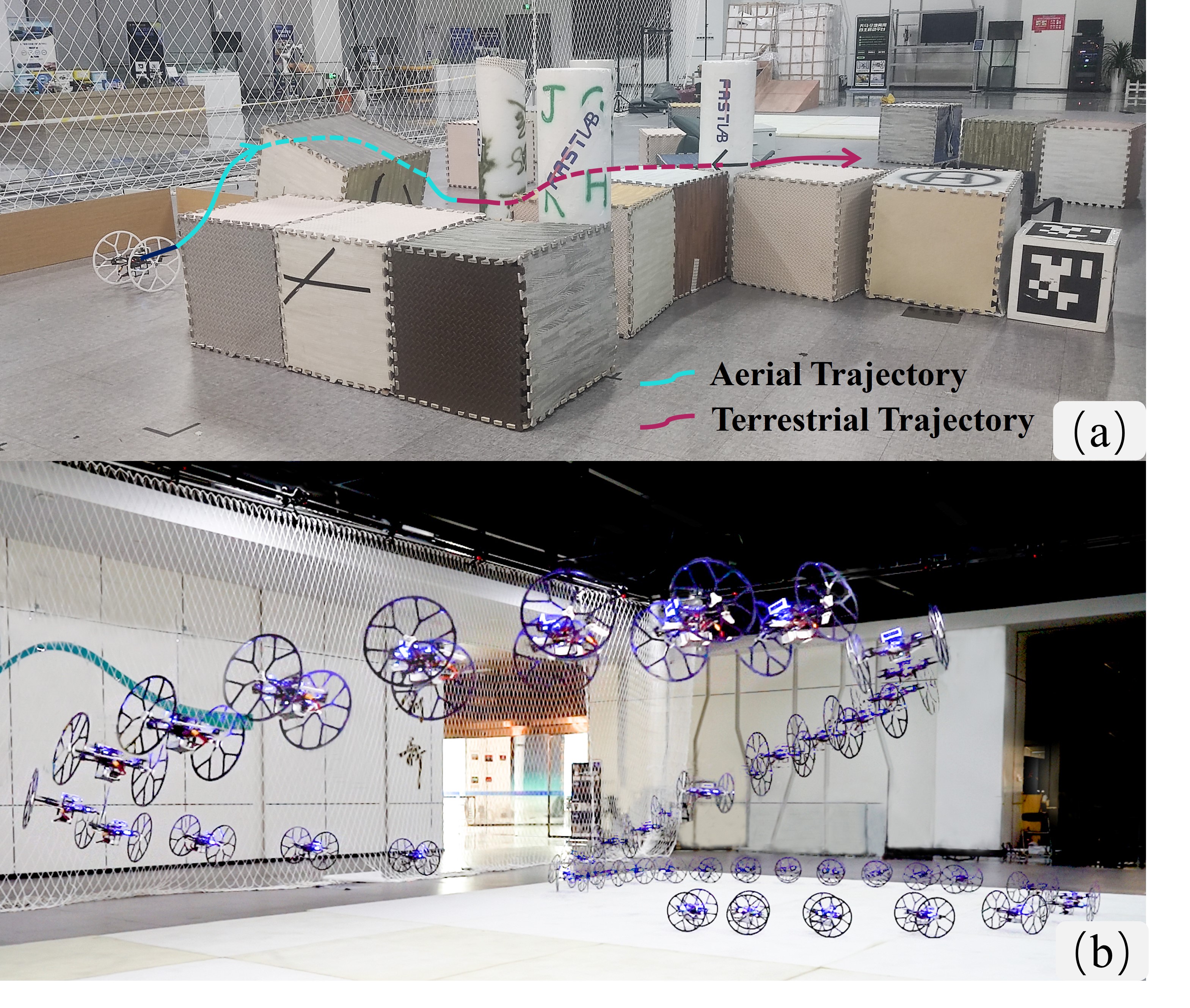}
	\captionsetup{font={small}}
	\caption{
		The real-world experiments. a) An autonomous navigation test in an unknown dense environment. b) A Terrestrial-aerial hybrid trajectory tracking test in which the maximal commanded velocity and acceleration reach $3m/s$ and $2.5m/s^2$, respectively.
	}
	\label{pic:experiments_scene}
	\vspace{-1.0cm}
	
	\end{figure}
   
	In our previous work\cite{zhang2022autonomous}, we propose a navigation framework for passive-wheeled TABV configurations (originally conceived by Kalantari et al. \cite{kalantari2013design}), which achieves state-of-the-art planning performance and controlling accuracy. However, our previous work is based on the dynamics and differential flatness of general quadrotors, while the dynamics of the TABV in terrestrial locomotion are overly simplified. As a result, the planning results are dynamically infeasible in certain cases. Furthermore, the control error considerably increases when tracking aggressive trajectories.  
	
\begin{figure}
	\centering
	\includegraphics[width=0.9\linewidth]{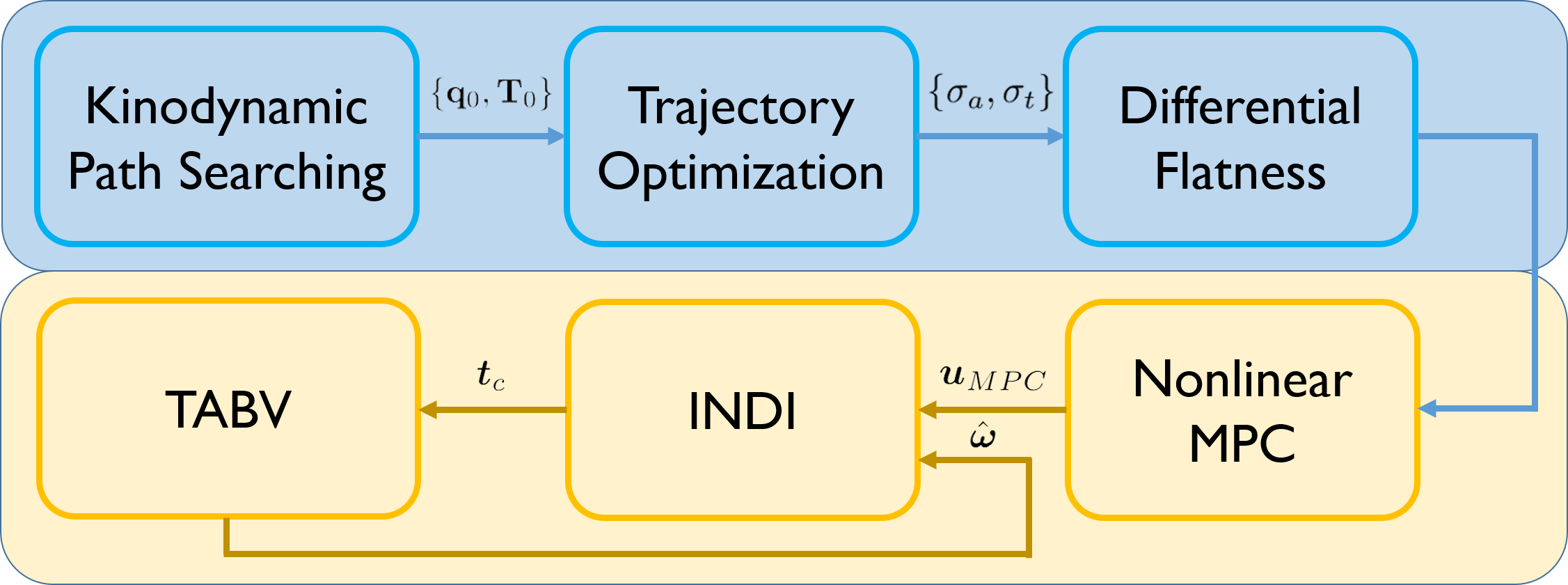}
	\captionsetup{font={small}}
	\caption{
		A diagram of the proposed planning and control framework.
	}
	\label{pic:software_architecture}
	\vspace{-1.1cm}
	
	\end{figure}

	To address the issues mentioned above, we propose a unified model-based planning and control framework for passive-wheeled TABVs, as illustrated in Fig.\ref{pic:software_architecture}. Chiefly, we model the nonlinear bimodal dynamics of passive-wheeled TABVs and their differential flatness in a unified manner. In trajectory planning, we present an optimization-based formulation utilizing the proposed TABV differential flatness map. It allows the complete states and constraints of both locomotion modes to be analytically derived from the flat output. This formulation simplifies the planning problem while ensuring dynamic feasibility. Furthermore, the unified modeling naturally guarantees continuity and smoothness during mode transitions. In motion control, we propose a bimodal tracking controller based on nonlinear model predictive control (NMPC), which exploits the full capability of the TABV without violating dynamic constraints. It tracks the reference state recovered from the planned flat output while satisfying the control limits and state constraints. Seamless mode transition is achieved thanks to proposed the unified modeling. We adopt incremental nonlinear dynamic inversion (INDI) as the inner-loop angular controller for robustification.


    We conduct extensive experiments and benchmark comparisons on a customized TABV platform in challenging real-world environments to verify the proposed methods. The results demonstrate that our method outperforms our previous work \cite{zhang2022autonomous} in planning quality and control performance. The contributions of this paper are as the followings:
    
\begin{itemize}
	\item [1)] 
	A unified dynamic model for a passive-wheeled TABV configuration, including the corresponding differential flatness that benefits trajectory planning and tracking control.
	\item [2)]
	A flatness-based TABV trajectory planner that guarantees dynamical feasibility, continuity, and smoothness for bimodal locomotion while maintaining real-time performance.
	\item [3)]
	An NMPC-based TABV tracking controller that achieves accurate and robust trajectory tracking and seamless mode transition without violating dynamic constraints.
	\item [4)]
	A set of real-world tests and benchmark comparisons that validate the the planning quality and control performance of the proposed methods.
	
\end{itemize}

	\section{Related Work} 
	\label{sec:related_works}	
 	\subsection{TABV Trajectory Planning}

	Many of the previous works on TABV trajectory planning aim to find a safe and efficient geometric path while considering the extra energy consumption in aerial locomotion \cite{araki2017multi, sharif2018energy, choudhury2019dynamic}. However, since no high-order information such as velocity or acceleration is available from a geometric path, these methods can hardly be applied to real-world scenarios. Fan et al. \cite{fan2019autonomous} propose a primitive-based local planner that generates minimum-snap trajectories under the guidance of geometric path searching. Nevertheless, this method does not take the terrestrial dynamics into account, thus cannot guarantee high-fidelity feasibility. In our previous work \cite{zhang2022autonomous}, we base trajectory planning on the differential flatness of general quadrotors and handle the nonholonomic dynamics in terrestrial locomotion by constraining the trajectory curvature. However, this indirect modeling bypasses terrestrial dynamics and cannot ensure physical feasibility as well. Wu et al. \cite{wu2023unified} propose a model-based planning approach that enables optimizing the full-state trajectories considering both terrestrial and aerial dynamics. The bimodal dynamics are modeled in a unified manner utilizing complementary constraints. However, this approach is time-consuming due to the integration of differential equations introduced by the nonlinear dynamics. In order to improve the computational efficiency, this method requires warm starting of the previously generated trajectories. However, as reported by Wu et al. \cite{wu2023unified}, when the newly observed obstacles intersect with the previously generated trajectory which results in the failure of the warm start mechanism, the computation time greatly increases and paused behaviors happen. To sum up, the  computation burden limits this approach's application in realistic settings. 
	
	In this work, we encode feasibility constraints in trajectory planning utilizing the proposed terrestrial-aerial bimodal differential flatness. The trajectory planning problem is transformed into optimizing continuous trajectories of the flat output and avoids explicit integration of system dynamics, thus ensuring both solution quality and computational efficiency.

	\subsection{TABV Tracking Control}
	Several approaches have been proposed to achieve bimodal trajectory tracking for passive-wheeled TABVs. A common approach is to use a cascaded control framework similar to a general quadrotor controller based on a simplified dynamic model \cite{colmenares2019nonlinear, fan2019autonomous, zhang2022autonomous, yang2022sytab}. In such a cascaded control architecture, the resulting control command is either too conservative or too aggressive because the nonlinear dynamics are not precisely captured\cite{sun2022comparative}. Moreover, the state and input limit cannot be properly handled. Atay et al.\cite{Atay2021ControlAC} derive the complete nonlinear dynamics and differential flatness of passive-wheeled TABVs and propose a model-based control system considering control allocation. However, the map from the flatness output to the system state is not considered, making this approach not applicable to trajectory tracking.
	
	In this work, we adopt NMPC, a control method that optimizes the behavior of a nonlinear system based on the system dynamics and constraints over a receding horizon, for TABV tracking control. Leveraging the unified dynamic model (as demonstrated in Sect.\ref{sec:modelling}), NMPC enables both accurate trajectory tracking and smooth locomotion mode transition. Moreover, the predictive nature of NMPC allows the TABV to react in advance to aggressive maneuvers that will occur in the near future. Although NMPC is more computationally demanding than non-predictive methods, previous works have shown that it can be applied to trajectory tracking control\cite{sun2022comparative, nan2022nonlinear} even using onboard computers with relatively low computing power.

	\section{PRELIMINARIES}
	\label{sec:modelling}
\begin{figure}
		\centering
		\includegraphics[width=0.8\linewidth]{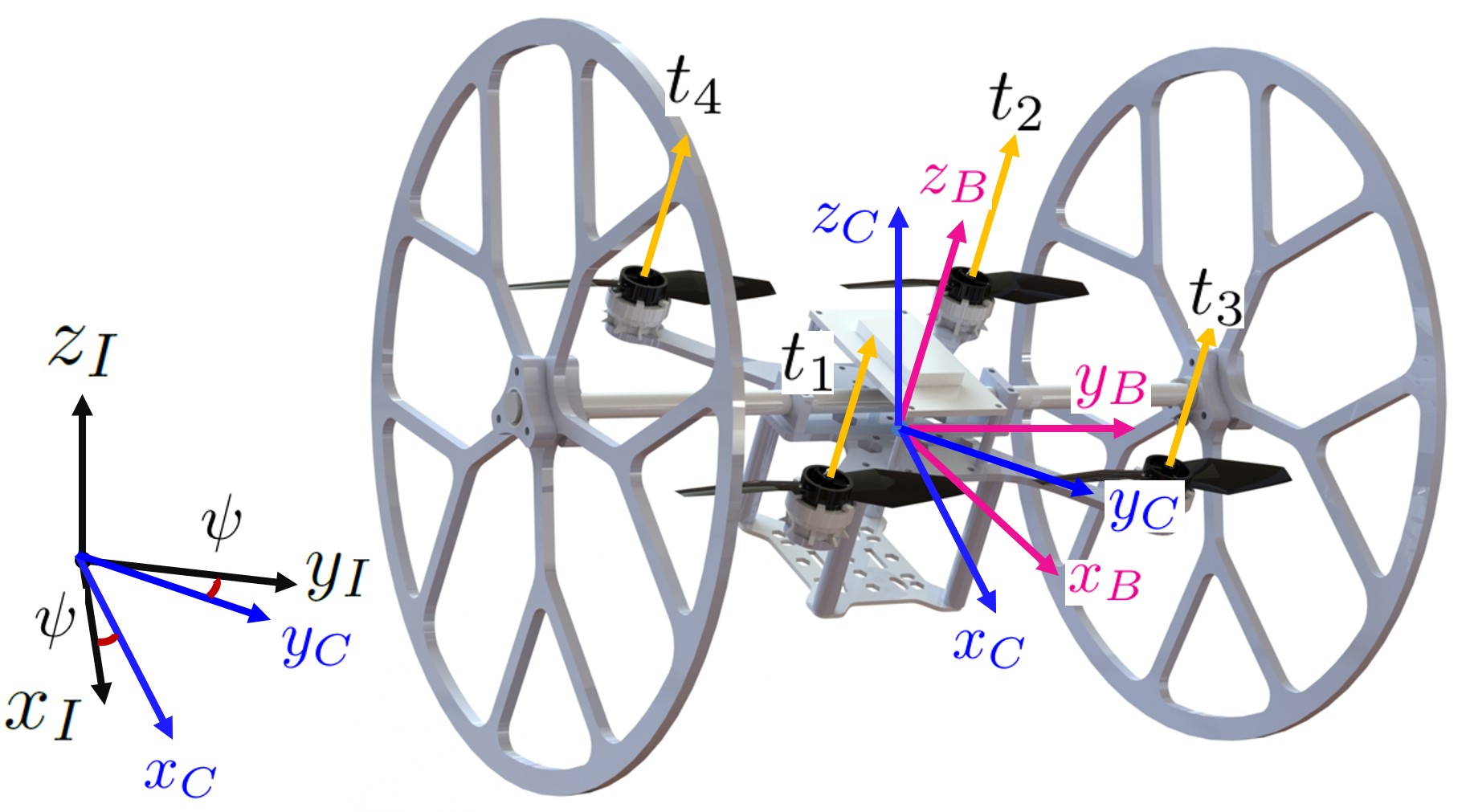}
		\captionsetup{font={small}}
		\caption{
			Illustration of the reference frames. Three frames are introduced: inertial frame ($\boldsymbol{x}_I - \boldsymbol{y}_I-\boldsymbol{z}_I$) with $\boldsymbol{Z}_I$ pointing in the opposite direction of the gravity vector, body frame ($\boldsymbol{x}_B - \boldsymbol{y}_B-\boldsymbol{z}_B)$) with $\boldsymbol{z}_B$ aligned with the rotor thrust vector, and intermediate frame ($\boldsymbol{x}_C - \boldsymbol{y}_C-\boldsymbol{z}_C$) which is separated from the inertial frame by the yaw rotation $\psi$. 
		}
		\label{pic:tabv_frame}
		\vspace{-1.1cm}
	\end{figure}

	In this section, we will introduce the unified TABV dynamics model and its corresponding differential flatness. The reference frames are defined in Fig.\ref{pic:tabv_frame}. 

	\subsection{Unified TABV Dynamics}
	\label{sec:dynamics}
	We define the state and input vectors of the TABV as 
	$\boldsymbol{x} = [
		\boldsymbol{p}_I^T, \boldsymbol{q}^T, \boldsymbol{v}_I^T, \boldsymbol{\omega}_B^T
	]^T$ and $\boldsymbol{u} = [T , \boldsymbol{\tau}_B^T ]^T$, respectively. Within the full state vector $\boldsymbol{x}$, $\boldsymbol{p}_I$ and $\boldsymbol{v}_I$ represent the three-dimensional position and velocity in the inertial frame. $\boldsymbol{q} = [ q_w, q_x, q_y, q_z]^T$ is the robot attitude expressed by unit quaternion. $\boldsymbol{\omega}_B = p\boldsymbol{x}_B + q\boldsymbol{y}_B + r\boldsymbol{z}_B$ is the three-dimensional angular velocity in the body frame. Within the input vector $\boldsymbol{u}$,  $T$ denotes the collective rotor thrust and $\boldsymbol{\tau}_B$ is the rotor torque expressed in the body frame. We model the TABV as a rigid body with total mass $m$ and inertia matrix $\boldsymbol{M}$. We consider the situation where the TABV moves on flat ground. The robot dynamics and motor dynamics can then be expressed as:
	\begin{equation}
		\label{eq:transdynamics} 
	m\ddot{\boldsymbol{p}_I}= T \boldsymbol{z}_B - mg\boldsymbol{z}_I + \boldsymbol{F}_e,
	\end{equation}
	\begin{equation}
	\label{eq:rotationdynamics} 
		\boldsymbol{M} \dot{\boldsymbol{\omega}_B} = \boldsymbol{\tau}_B-\boldsymbol{\omega}_B \times \boldsymbol{M} \boldsymbol{\omega}_B + \boldsymbol{\tau}_e ,
	\end{equation}
	\begin{equation}
	\label{eq:motordynamics} 
	\left[\begin{array}{l}T \\ \boldsymbol{\tau}_B\end{array}\right]
	\! = \! \mathcal{M} \boldsymbol{t}
	\setlength{\arraycolsep}{1.5pt}
	\! = \!\left[\begin{array}{cccc}1 & 1 & 1 & 1 \\ -L / \sqrt{2}  & L / \sqrt{2}  & L / \sqrt{2}  & -L / \sqrt{2}  \\ -L / \sqrt{2}  & L / \sqrt{2}  & -L / \sqrt{2}  & L / \sqrt{2} \\ -c_m / c_t & -c_m / c_t & c_m / c_t & c_m / c_t\end{array}\right] \boldsymbol{t},
	\end{equation}
	where $g = 9.81 m/s^2 $ is the magnitude of gravity, $\boldsymbol{F}_e$ is the external force, $\boldsymbol{\tau}_e$ is the external torque. $\boldsymbol{t} = [t_1, t_2, t_3, t_4]^T$ is the rotor thrust vector, $\mathcal{M}$ is the allocation matrix of a standard $X-$shape configuration, in which $L$ is the arm length, $c_m$ and $c_t$ are the rotor torque and thrust coefficient, respectively.

	Note that (\ref{eq:transdynamics}-\ref{eq:motordynamics}) are applicable to both terrestrial and aerial modes because the TABV is driven by rotor thrusts in both modes. The only difference 
	lies in the additional contact forces from the ground in terrestrial mode. To be specific, $\boldsymbol{F}_e = \boldsymbol{F}_a$ in aerial mode, while $\boldsymbol{F}_e = \boldsymbol{F}_a + \boldsymbol{F}_c$ in terrestrial mode, where $\boldsymbol{F}_a$ is the exogenous aerodynamic force (rotor drag for e.g.), $\boldsymbol{F}_c = \boldsymbol{q_}\psi \odot  [F_d, F_l, F_n]^T$ is the ground contact force that ($\boldsymbol{q_}\psi$ denotes the quaternion that only includes yaw rotation). Among the components of $\boldsymbol{F}_c$, $F_d$ is the rolling resistance, $F_n$ is the normal force, $F_l$ is the lateral force that prevents side slipping. The corresponding nonholonomic constraint is written as:
	\begin{equation}	
		(\boldsymbol{q}^{-1} \odot \boldsymbol{v}_I) \cdot \boldsymbol{e}_2 = 0,
	\end{equation}
	where $\boldsymbol{e}_2 = [0, 1, 0]^T$, $\odot$ denotes the rotation of a vector using a quaternion. In addition, because of the normal force, the TABV keeps in contact with the ground:
	\begin{equation}	
		\boldsymbol{p}_I \cdot \boldsymbol{e}_3 = 0,
	\end{equation}
	where $\boldsymbol{e}_3 = [0, 0, 1]^T$.
	
	As the effect of the aerodynamic force only becomes noticeable during highly aggressive locomotion, we neglect ${F}_a$ in this work. As long as the wheel is in contact with the ground, the rolling friction force $F_d$ is always present and almost maintains a small constant value. Therefore, we omit $F_d$ as well. The external torque $\boldsymbol{\tau}_e$ in both modes is identical, and we leave out this term because it is hard to model in real-world environments\cite{sun2022comparative}. It is compensated later by the INDI inner-loop controller (introduced in Sect.\ref{sec:INDI}). 
	
	Finally, the system dynamics $f(\boldsymbol{x}, \boldsymbol{u})$ are given by: 
	
	\begin{subequations}
	\begin{align}
	\dot{\boldsymbol{x}}  =\left[\begin{array}{c}
		\dot{\boldsymbol{p}_I} \\
		\dot{\boldsymbol{q}} \\
		\dot{\boldsymbol{v}_I} \\
		\dot{\boldsymbol{\omega}_B}
		\end{array}\right] & = \left[\begin{array}{c}
		\boldsymbol{v}_I \\
		\frac{1}{2} \boldsymbol{q} \circ\left[\begin{array}{l}
			0 \\
			\boldsymbol{\omega}_B
		\end{array}\right] \\
		(\boldsymbol{F}_n + T\boldsymbol{z}_B ) / m +\boldsymbol{g} \\
		\boldsymbol{M}^{-1}\left[\boldsymbol{\tau}_B-\boldsymbol{\omega}_B \times \boldsymbol{M} \boldsymbol{\omega}_B \right]
	\end{array}\right] \label{eq:dynamics_new} \\
		\text { s.t. } 	&F_n[(\boldsymbol{q}^{-1} \odot \boldsymbol{v}_I) \cdot \boldsymbol{e}_2)] = 0, \label{eq:nonholoconstraint}\\
		&F_n(\boldsymbol{p}_I \cdot \boldsymbol{e}_3) = 0,  \label{eq:groundconstraint}
 	\end{align}
	\end{subequations}	
	where $\boldsymbol{F}_n = [0, 0, F_n]^T$. Since $F_n$ is equal to zero in aerial mode,  (\ref{eq:dynamics_new}-\ref{eq:groundconstraint}) hold in both locomotion modes.  

	\subsection{Differential Flatness}
	\label{sec:flatness}
	This section introduces the differential flatness of TABV bimodal dynamics. In aerial locomotion, a TABV is no different from a general quadrotor whose differential flatness has been well studied by previous works\cite{MelKum1105, faessler2017differential, watterson2019control, wang2022robust}. The position and yaw are typically taken as the flat output, denoted as $\sigma_a=[\boldsymbol{p}_{I, x}, \boldsymbol{p}_{I, y}, \boldsymbol{p}_{I, z}, \psi]^T$. In practice, yaw planning is independent of the three-dimensional position planning and does not affect the quadrotor's translational motion. 
	
	Regarding terrestrial locomotion, we demonstrate that the TABV dynamics remain differentially flat with the flat output given by 
	\begin{equation}
	\sigma_t=[\boldsymbol{p}_{I, x}, \boldsymbol{p}_{I, y}, \boldsymbol{p}_{I, z}, \theta]^T,
	\end{equation}	
	where $\theta$ is the pitch. The derivation of the differential flatness draws on the work of Mellinger et. al.\cite{MelKum1105}.
	
	First, the three-dimensional position ($\boldsymbol{p}_{I}$), velocity ($\boldsymbol{v}_{I}$), and acceleration ($\boldsymbol{a}_{I}$) are the first three terms of $\sigma_t, \dot{\sigma}_t,$ and $ \ddot{\sigma}_t$, respectively. Afterward, we show that the attitude, body angular velocity, and control input can also be derived from $\sigma_t$ and its derivatives. 

	Since the heading of the TABV is parallel with the direction of $\boldsymbol{v}_I$, it can be expressed as
	\begin{equation}
		\label{eq:yaw_equation}
		\psi = arctan2\left(\eta\dot{\boldsymbol{p}}_{I,y}, \eta\dot{\boldsymbol{p}}_{I,x}\right), 
	\end{equation}
	where $\eta \in\{-1,1\}$ is the motion direction of TABV ($\eta = 1$ means forward moving and $\eta = -1$ means backward moving). Then we can write the unit vector $\boldsymbol{x}_C=\left[\cos \psi, \sin \psi, 0\right]^T$ as shown in Fig.\ref{pic:tabv_frame}.

%
%
%
	
	Because the horizontal components of $\boldsymbol{F}_n$ are not considered, left multiplying (\ref{eq:transdynamics}) by $\boldsymbol{x}_C$ gives:
	\begin{equation}
	\label{eq:T_formulation}
T = \frac {m\ddot{\boldsymbol{p}}_{I} \cdot \boldsymbol{x}_C} {  sin\theta_e}. 
	\end{equation}


	Thus, the collective thrust $T$ is recovered from the flat output.
	
	As for the attitude, note that no roll motion occurs since we assume that the TABV moves on flat ground, then $\boldsymbol{z}_B$ can be determined as follows:

	 \begin{equation}
	\begin{aligned} 	\boldsymbol{z}_B & =\frac{\mathbf{t}}{\|\mathbf{t}\|}, \mathbf{t}=\left[\boldsymbol{a}_l \cos \psi, \boldsymbol{a}_l \sin \psi, \sqrt{T^2 - \boldsymbol{a}_l^2}\right]^T , \\
			\boldsymbol{a}_l & = (\boldsymbol{v}_{I,x} \boldsymbol{a}_{I,x} + \boldsymbol{v}_{I,y} \boldsymbol{a}_{I,y})) / \sqrt{\boldsymbol{v}_{I,x}^2 + \boldsymbol{v}_{I,y}^2}, 
	\end{aligned}
	\end{equation}

	where $\boldsymbol{a}_l$ is the longitudinal acceleration.
	Then, $\boldsymbol{x}_B$, $\boldsymbol{y}_B$, and the attitude ${ }^I R_B$ (expressed by rotation matrix here) can be calculated by
	\begin{equation}
		\begin{aligned}
\boldsymbol{y}_B & =\frac{\boldsymbol{z}_B \times \boldsymbol{x}_C}{\left\|\boldsymbol{z}_B \times \boldsymbol{x}_C\right\|}, \boldsymbol{x}_B=\boldsymbol{y}_B \times \boldsymbol{z}_B, \\
 { }^I R_B & = \left[\begin{array}{lll}
	\boldsymbol{x}_B & \boldsymbol{y}_B & \boldsymbol{z}_B
\end{array}\right].
\end{aligned}
	\end{equation}	
	 Therefore, the attitude of TABV can be determined from $\sigma_t$ and its derivatives as well. 
 	
	We then take the first derivative of (\ref{eq:transdynamics}):

	\begin{equation}
	\label{eq:transdynamics_derivative} 
	m\dddot{\boldsymbol{p}_I}= \dot{T}\boldsymbol{z}_B + \boldsymbol{\omega}_B \times T\boldsymbol{z}_B + \dot{F}_n\boldsymbol{z}_I.
	\end{equation}	
	
	We define the vector $\mathbf{h}_\omega = \boldsymbol{\omega}_B \times \boldsymbol{z}_B$ as
	\begin{equation}
	\label{eq:transdynamics_derivative} 
	\mathbf{h}_\omega = \frac {1} {T} \left(m\dddot{\boldsymbol{p}_I} - \dot{T}\boldsymbol{z}_B - \dot{F}_e\boldsymbol{z}_I\right) = f(\sigma_t, \dot{\sigma_t}, \ddot{\sigma_t}, \dddot{\sigma_t}).
	\end{equation}	 	
	
	Afterwards, the components of $\omega_{B}$ are found as 
	\begin{equation}
	p=-\mathbf{h}_\omega \cdot \mathbf{y}_B, 
	q=\mathbf{h}_\omega \cdot \mathbf{x}_B,
	r = \dot{\psi} \mathbf{z}_I \cdot \mathbf{z}_B.
	\end{equation}	

	Finally, the rotor torque $\boldsymbol{\tau}_B$ can be calculated from (\ref{eq:rotationdynamics}) since the other terms in (\ref{eq:rotationdynamics}) are functions of $\sigma_t$ and its derivatives.
	
	We find that $\boldsymbol{p}_I$ exists in both $\sigma_a$ and $\sigma_t$. In trajectory planning, $\boldsymbol{p}_I$ is planned at first. Then, $\psi$ is set to be parallel with the direction of $\boldsymbol{v}_I$ in aerial locomotion as well for the continuity of trajectories. In terrestrial locomotion, $\theta$ is calculated from (\ref{eq:T_formulation}) in which the reference thrust is set to be constant.

	\section{TABV Trajectory Planning}
	\label{sec:planning}
	In this section, we elaborate on the proposed flatness-based planner. Utilizing the differential flatness derived in Sect.\ref{sec:flatness}, the trajectory planning problem can be reduced to planning the three-dimensional position. We follow the typical hierarchical framework with a kinodynamic path searching front-end and a spatial-temporal trajectory optimization back-end.
	
	\subsection{Kinodynamic Path Searching}
	Kinodynamic path searching method aims at finding a safe and dynamically feasible path based on a simplified dynamic model. We adopt the hybrid-state A* algorithm\cite{dolgov2008practical} for path searching. An energy cost is exerted on aerial motion primitives to penalize flying locomotion. We refer the readers to our previous works\cite{zhang2022autonomous} for more details. To deal with the nonholonomic dynamics in terrestrial locomotion which is neglected in our previous work, we use the following differential wheeled robot dynamics for terrestrial motion primitives generation.
	\begin{equation}
	\left\{
	\begin{aligned}
		{\dot p_x}& = v \cos\varphi, \\
		{\dot p_y}& = v \sin \varphi,  \\
		\dot{v}& = a, \\
		\dot{\varphi}& =  \omega,\\
	\end{aligned}
	\right.
	\end{equation}

	where the state vector includes two-dimensional position $[p_x, p_y]^T$, longitudinal velocity $v$, and heading angle $\varphi$. Longitudinal acceleration $a$ and heading angular velocity $\Omega$ are taken as the control input, i.e., $\boldsymbol{u}_t = [v, \varphi]^T$. This modeling method prohibits lateral movement, which accords with the nonholonomic dynamics of TABV.
	
	In terms of aerial locomotion, the TABV dynamics can be simplified into a linear system written as 
	\begin{equation}
	\left[\begin{array}{c}
		\dot{\boldsymbol{p}} \\
		\dot{\boldsymbol{v}} \\
	\end{array}\right]=
	\left[\begin{array}{cc}
		\boldsymbol{0} & \boldsymbol{1} \\
		\boldsymbol{0} & \boldsymbol{0} \\
	\end{array}\right]\left[\begin{array}{l}
		\boldsymbol{p} \\
		\boldsymbol{v}
	\end{array}\right] + \left[\begin{array}{c}
	\boldsymbol{0} \\
	\boldsymbol{1} \\
	\end{array}\right] \boldsymbol{a},
	\end{equation}
	where $\boldsymbol{p}, \boldsymbol{v}$, and $\boldsymbol{a}$ are the three-dimensional position, velocity, and acceleration, respectively. $\boldsymbol{u}_a = \boldsymbol{a}$ is taken as the control input.  $\boldsymbol{u}_t$ and $\boldsymbol{u}_a$ is sampled simultaneously to generate both terrestrial and aerial motion primitives, leading to a terrestrial-aerial path that serves as the initial guess for trajectory optimization. 
	\subsection{Optimization Problem Formulation}
	In trajectory optimization, we adopt MINCO trajectory class\cite{wang2022geometrically} to generate minimum control effort spatial-temporal trajectories. It achieves linear-complexity spatial-temporal deformation of the flat-output trajectories. The full state defined in (\ref{eq:dynamics_new}) can then be recovered from the flat output. For an $m$-dimensional $M$-piece polynomial trajectory with degree $N = 2s - 1$, where $s$ is the order of the integrater chain (we choose $s = 3$ in this work), a linear-complexity map is constructed as
	\begin{equation}
	\mathbf{c}=\mathcal{M}(\mathbf{q}, \mathbf{T}),
	\end{equation}
	where $\mathbf{c}=\left(\mathbf{c}_1^T, \ldots, \mathbf{c}_M^T\right)^T \in \mathbb{R}^{2 M s \times m}$ is the polynomial coefficient matrix and $\mathcal{M}(\mathbf{q}, \mathbf{T})$ is the smooth map from intermediate waypoints $\mathbf{q}=\left(q_1, \cdots, q_{M-1}\right)^T \in \mathbb{R}^{m \times(M-1)}$ and a time allocation $\mathbf{T}=\left(T_1, \cdots, T_M\right)^T \in \mathbb{R}_{>0}^M$ to $\mathbf{c}$. The optimization variables are then converted to $\{\mathbf{q}, \mathbf{T}\}$ through $\mathcal{M}(\mathbf{q}, \mathbf{T})$. The $i$-th trajectory piece $p_i(t)$ can be expressed as
	\begin{equation}
	p(t)=p_i\left(t-t_{i-1}\right), \forall t \in\left[t_{i-1}, t_i\right],
	\end{equation}
	\begin{equation}
	p_i(t)=\mathbf{c}_i^{\mathrm{T}} \beta(t), \quad \forall t \in\left[0, T_i\right],
	\end{equation}
	where  $\beta(t):=\left[1, t, \cdots, t^N\right]^{\mathrm{T}} $ is the natural basis. We formulate the TABV trajectory generation problem as unconstrained nonlinear programming written as
	\begin{equation}
	\min _{\mathbf{q}, \mathbf{T}}\left[J_t, J_s, J_c, J_{n}\right] \cdot \mathbf{\lambda}, 
	\end{equation}

	where ${\lambda}$ is a weighting vector to trade off each cost function. In trajectory optimization, we sample three-dimensional constraint points on each trajectory piece denoted as $\tilde{\boldsymbol{p}}_{i, j}=p_i\left(\left(j / \kappa_i\right) \cdot T_i\right), j=0,1, \ldots, \kappa_i-1$ for discretizing the above cost functions. From path searching, the initial guess $\mathbf{q}_0$ and $\mathbf{T}_0$ is obtained, along with each piece's locomotion mode vector denoted as $\mathbf{l}=\left(l_1, \cdots, l_{M-1}\right)^T$, in which $l_i = 1$ means terrestrial locomotion and $l_i = 0$ means aerial locomotion.

	\subsubsection{Cost of Total time $J_t$}
	To accelerate the navigation progress, we first minimize the total time through $J_t=\sum_{i=1}^M T_i$.
	\subsubsection{Cost of State Limit $J_s$}
	We then limit the magnitude of linear velocity and acceleration to ensure the feasibility. The cost function is formulated as follows:
	\begin{equation}
	J_s = J_{s, v} + J_{s, a},
	\end{equation}
	\begin{equation}
	J_{s, v}= \sum_{i=1}^M \sum_{j=0}^{\kappa_i-1} \max \left\{\left\|\dot{\tilde{\boldsymbol{p}}}_{i,j}\right\|^2_2-v_{max}^2, 0\right\},
	\end{equation}
	\begin{equation}
	J_{s, a}=\sum_{i=1}^M \sum_{j=0}^{\kappa_i-1}\max \left\{\left\|\ddot{\tilde{\boldsymbol{p}}}_{i,j}\right\|^2_2 -a_{max}^2, 0\right\},
	\end{equation}
	where $v_{max}$ and $a_{max}$ are the linear velocity and acceleration thresholds, respectively. Note that $J_s$ is applicable to both terrestrial and aerial locomotion by confining the height of $\tilde{p}_{i,j}$ to be constant if $l_i = 1$.

	\subsubsection{Cost of Obstacle Avoidance  $J_c$} 
	To ensure a collision-free trajectory is generated, we use Euclidean Sign Distance Field (ESDF) map to obtain the distance and the corresponding gradient information of adjacent obstacles. The collision avoidance cost function $J_c$ is defined as follows:
	\begin{equation}
		J_c = \sum_{i=1}^M \sum_{j=0}^{\kappa_i-1} \max \left\{\left\|d_s - \bm{\mathcal{E}}({\tilde{\boldsymbol{p}}}_{i,j})\right\|_1, 0\right\}, 
	\end{equation}
	where $ {\bm{\mathcal{E}}({\cdot})} $ is the distance between TABV and the nearest obstacle obtained from ESDF, $d_s$ is the safety distance threshold.

    \subsubsection{Cost of Nonholonomic Dynamics $J_n$}
    In terrestrial locomotion, the TABV's velocity is limited to be parallel with the yaw heading due to the nonholonomic dynamics. To ensure this in trajectory optimization, we first align the directions of velocities with the desired heading angles in both initial and final states, then limit the heading angular velocity and acceleration to further guarantee dynamical feasibility along the whole trajectory. According to (\ref{eq:yaw_equation}), the heading angular velocity and acceleration can be expressed as 
    \begin{equation}
	\dot{\varphi}(\boldsymbol{p}) = \eta\frac{\ddot{\boldsymbol{p}}^{T} \bm{{\rm B}} \dot{\boldsymbol{p}}}{\dot{\boldsymbol{p}}^{T} \dot{\boldsymbol{p}}}, \bm{{\rm B}} = \begin{bmatrix}	0 & -1\\1 & 0\end{bmatrix},
	\end{equation}
    \begin{equation}
	\ddot{\varphi}(\boldsymbol{p}) = \eta \left( \frac{\dddot{\boldsymbol{p}}^{T} \bm{{\rm B}} \dot{\boldsymbol{p}}}{\dot{\boldsymbol{p}}^{T} \dot{\boldsymbol{p}}} - \frac{2\ddot{\boldsymbol{p}}^{T} \bm{{\rm B}} \dot{\boldsymbol{p}} \ddot{\boldsymbol{p}}^{T} \dot{\boldsymbol{p}} }{(\dot{\boldsymbol{p}}^{T} \dot{\boldsymbol{p}})^2}\right),
	\end{equation}	
	where $\boldsymbol{p}$ is short for $\tilde{\boldsymbol{p}}_{i,j}$. In practice, we only consider $\eta = 1$ because the customized TABV platform uses a limited-FOV stereo camera for sensing (details are available in Sect.\ref{sec:implementation_details}), and backward movements are not safe. The cost of nonholonomic dynamics $J_n$ is defined as
	\begin{equation}
	J_n = J_{s, \Omega} + J_{s, \alpha},
	\end{equation}
	\begin{equation}
		J_{s, \Omega}= \sum_{i=1}^M \sum_{j=0}^{\kappa_i-1}  l_i \cdot \max \left\{\left\|\dot{\varphi}({\tilde{\boldsymbol{p}}}_{i,j})\right\|^2_2-\Omega_{max}^2, 0\right\},
		\label{eq:omegacost}
	\end{equation}
	\begin{equation}
		J_{s, \alpha}= \sum_{i=1}^M \sum_{j=0}^{\kappa_i-1}  l_i \cdot \max \left\{\left\|\ddot{\varphi}({\tilde{\boldsymbol{p}}}_{i,j})\right\|^2_2-\alpha_{max}^2, 0\right\},
		\label{eq:alphacost}
	\end{equation}
	where $\Omega_{max}$ and $\alpha_{max}$ are the heading angular velocity and acceleration threshold, respectively. In (\ref{eq:omegacost}) and (\ref{eq:alphacost}), if $l_i = 0$, the corresponding aerial trajectory piece is not affected by this cost.
	
	After the final trajectory is generated, we sample points and their high-order derivatives along the trajectory and restore the desired full state and input from the differential flatness derived in Sect.\ref{sec:flatness} as the reference for trajectory tracking control.

	\section{TABV Trajectory Tracking Control}
	\subsection{Nonlinear Model Predictive Control Formulation}
	\label{sec:NMPC} 
	Given the TABV system dynamics $f(\boldsymbol{x}, \boldsymbol{u})$ (\ref{eq:dynamics_new}) and the reference trajectories generated by the TABV trajectory planner, NMPC finds the control commands by solving an optimal control problem (OCP) in a receding horizon manner. The NMPC formulation minimizes the error between predicted states and reference states in a finite look-ahead time horizon $[t_0, t_0 + h]$. For numerical calculation, the formulation is discretized with a fixed time step $dt$ and the corresponding horizon length $N = h / dt$. Then, the NMPC problem is transcribed
	into constrained nonlinear programming that generates the optimal control command sequence $\mathbf{U}^* \in \mathbb{R}^{4 \times N}$: 
	
	 \begin{equation}
	\begin{aligned} &\mathbf{U}^* \!= \! \operatorname{argmin}_{\boldsymbol{u}} \sum_{i=k}^{k+N-1}\left(\tilde{\boldsymbol{x}}_i^T \boldsymbol{W}_x \tilde{\boldsymbol{x}}_i+\tilde{\boldsymbol{u}}_i^T \boldsymbol{W}_u \tilde{\boldsymbol{u}}_i\right)+\tilde{\boldsymbol{x}}_N^T \boldsymbol{W}_x \tilde{\boldsymbol{x}}_N, \\  &\text { s.t. } \quad  \boldsymbol{x}_{i+1}=f\left(\boldsymbol{x}_i, \boldsymbol{u}_i\right),  \boldsymbol{u}_{min}<\boldsymbol{u}_i<\boldsymbol{u}_{max}, \end{aligned}
 	\end{equation}
 	where $k$ is the current time step, $\tilde{\boldsymbol{x}}_i = \boldsymbol{x}_{ref, i} - \boldsymbol{x}_{i}$ and $\tilde{\boldsymbol{u}}_i = \boldsymbol{u}_{ref, i} - \boldsymbol{u}_{i}$ are the state and input error, respectively. $\tilde{\boldsymbol{x}}_N = \boldsymbol{x}_{ref, N} - \boldsymbol{x}_{N}$ is the end state error. $\boldsymbol{W}_x$ and $\boldsymbol{W}_u$ are the state and input weight matrix written as:
 	\begin{equation}	
 	\begin{aligned}
 	&\boldsymbol{W}_x=\textrm{diag}\left(\left[\begin{array}{llll}
 		\boldsymbol{W}_{x, p} & \boldsymbol{W}_{x, q} & \boldsymbol{W}_{x, v} & \boldsymbol{W}_{x, \omega} 
 	\end{array}\right]\right), \\
 	&\boldsymbol{W}_u=\textrm{diag}\left(\left[\begin{array}{lll}
 		W_{u, T} & \boldsymbol{W}_{u, \tau_{xy}}  & W_{u, \tau_z}
 	\end{array}\right]\right).
 	\end{aligned}
 	\end{equation}
 	
 	To combine both terrestrial and aerial dynamics, We define $\mu_g \in \{0, 1\} \in \mathbb{N}$ and introduce two equivalent constraints as follows: 
	\begin{equation}	
		\mu_g[(\boldsymbol{q}^{-1} \odot \boldsymbol{v}_I) \cdot \boldsymbol{e}_2] = 0,
	\end{equation}
	\begin{equation}	
		\mu_g(\boldsymbol{p}_I \cdot \boldsymbol{e}_3) = 0,
	\end{equation}
	where $\mu_g = 0$ for aerial mode and $\mu_g = 1$ for terrestrial mode.
	
	\subsection{Incremental Nonlinear Dynamic Inversion}
	\label{sec:INDI} 
	As mentioned in Sect.\ref{sec:dynamics}, the external torque $\boldsymbol{\tau}_e$ is omitted in TABV dynamics. As validated in recent works\cite{sun2022comparative}, INDI reveals distinct advantages for serving as a low-level controller by virtue of its simplicity and robustness. To be specific, INDI estimates $\boldsymbol{\tau}_e$ from instantaneous sensor measurements rather than a precise dynamic model. We adopt the INDI version proposed by Sun et al.\cite{sun2022comparative}. According to the robot rotational dynamics (\ref{eq:rotationdynamics}), we can estimate $\boldsymbol{\tau}_e$ as:
	\begin{equation}
	\label{eq:indi_1}
	\boldsymbol{\tau}_e=\boldsymbol{M} \dot{\hat{\boldsymbol{\omega}}}-\hat{\boldsymbol{\tau}}+\hat{\boldsymbol{\omega}} \times \boldsymbol{M} \hat{\boldsymbol{\omega}}, 
	\end{equation}
	where the superscript  $\hat{}$ means the corresponding term is obtained from sensor measurements and then low-pass filtered with the same cut-off frequency. Angular acceleration $\dot{\hat{\boldsymbol{\omega}}}$ is obtained from numerical differentiation of the measured angular rate. Assuming that $\boldsymbol{\tau}_e$ is slow-changing compared with the low-pass filter dynamics, the delay of $\boldsymbol{\tau}_e$ is negligible \cite{sun2022comparative}. Then, substitution of (\ref{eq:indi_1}) into (\ref{eq:rotationdynamics}) gives:
	
	\begin{equation}	
		\label{eq:indi_dynamics}
		\begin{aligned}
			\boldsymbol{M}\dot{\boldsymbol{\omega}}_B & 
			 = (\boldsymbol{\tau}_B - \boldsymbol{\omega}_B \times \boldsymbol{M} \boldsymbol{\omega}_B) + (\boldsymbol{M} \dot{\hat{\boldsymbol{\omega}}}-\hat{\boldsymbol{\tau}}+\hat{\boldsymbol{\omega}} \times \boldsymbol{M} \hat{\boldsymbol{\omega}}) \\
			& \approx  (\boldsymbol{\tau}_B - \hat{\boldsymbol{\tau}}) + \boldsymbol{M} \dot{\hat{\boldsymbol{\omega}}}.
		\end{aligned}
	\end{equation}

 	With the desired control input  $\boldsymbol{u}_{MPC} = [T , \boldsymbol{\tau}_B^T ]^T$ solved by NMPC, we can acquire the desired angular acceleration $\ddot{\boldsymbol{\omega}}_{B}^d$from (\ref{eq:dynamics_new}):
	\begin{equation}	
		\label{eq:nmpc_results}
	\ddot{\boldsymbol{\omega}}_{B}^d = \boldsymbol{M}^{-1}[\boldsymbol{\tau}_{B} - \boldsymbol{\omega} \times \boldsymbol{M} \boldsymbol{\omega}].
	\end{equation}
 
	From (\ref{eq:indi_dynamics}) and (\ref{eq:nmpc_results}), the commanded body torque $\boldsymbol{\tau}_{B}^d$ can be obtained as: 
	\begin{equation}	
		\label{eq:indi_result}
		\boldsymbol{\tau}_{B}^d = \hat{\boldsymbol{\tau}} + \boldsymbol{M}(\dot{\boldsymbol{\omega}}_{B}^d -  \dot{\hat{\boldsymbol{\omega}}}). 
	\end{equation}

	Finally, the commanded rotor thrust vector $\boldsymbol{t}_c$ can be calculated from (\ref{eq:motordynamics}) for low-level motor control.

	\section{Results}
	\label{sec:results}
	
	\begin{figure}
		\centering
		\includegraphics[width=0.8\linewidth]{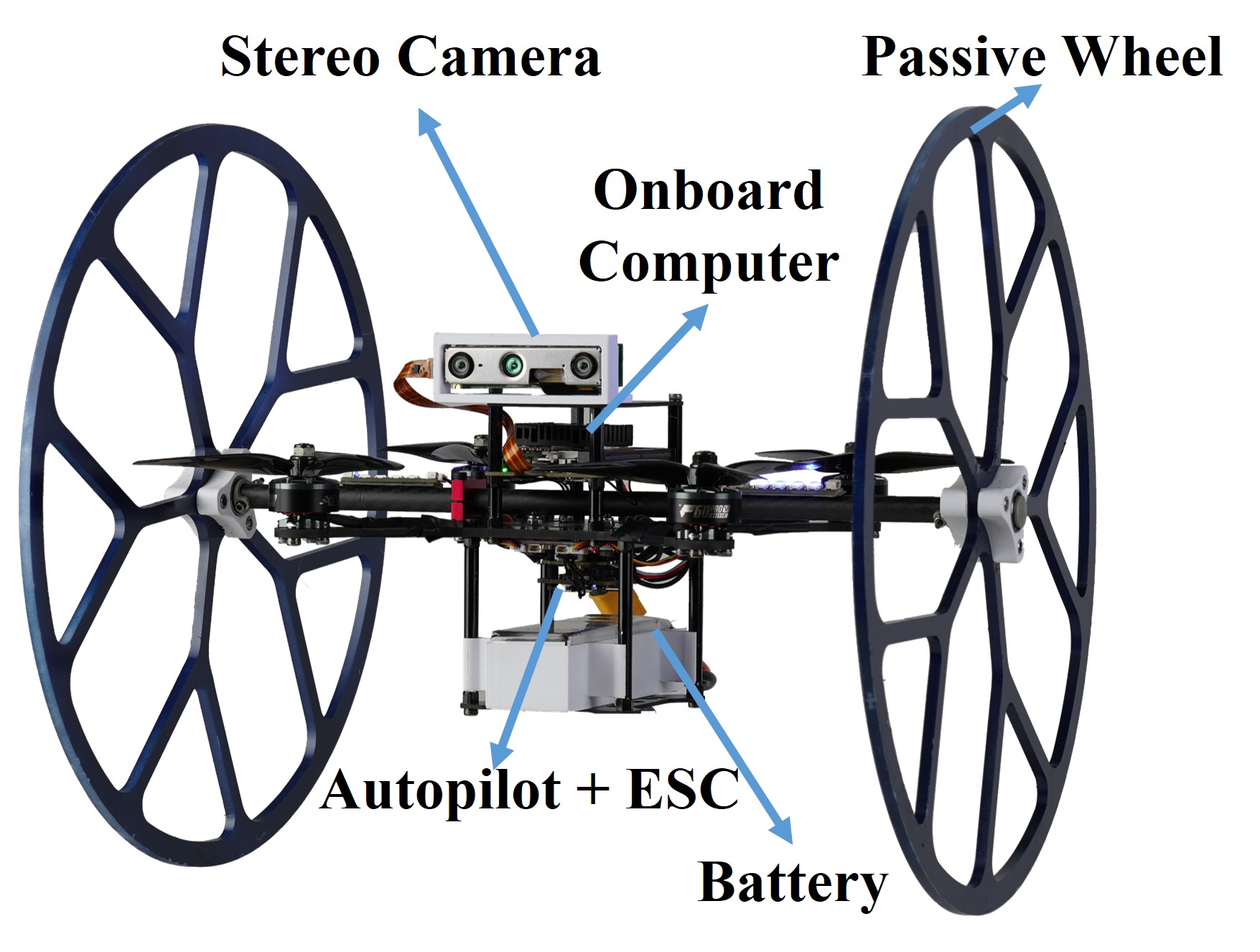}
		\captionsetup{font={small}}
		\caption{
			Illustration of the customized TABV platform.
		}
		\label{pic:tabv_fig_new}
		\vspace{-0.6cm}
	\end{figure}

	\subsection{Implementation Details}
\label{sec:implementation_details}
	To demonstrate and verify the proposed approaches in real-world environments, we customize a TABV platform, as shown in Fig.\ref{pic:tabv_fig_new}. The corresponding physical parameters are listed in Tab.\ref{tab:physical_parameters}. The 6D localization of the TABV and the environmental information  are obtained from a NOKOV motion capture system\footnote{https://www.nokov.com} and a Realsense D430 stereo camera, respectively. The mapping, planning, and control algorithms run on a Jetson Xavier NX onboard computer. In motion planning, the proposed  unconstrained optimization problem is solved by 
	LBFGS-Lite\footnote{https://github.com/ZJU-FAST-Lab/LBFGS-Lite}. The weighting vector ${\lambda}$ is set to be $[5, 6, 100, 5]^T$ in experiments. In motion control, the NMPC problem is solved by ACADO\cite{houska2011acado} with qpOASES\cite{ferreau2014qpoases} at 200 Hz. The horizon length $N$ and the time step $dt$ is set to be $20$ and $70ms$, respectively. The NMPC weights are shown in Tab.\ref{tab:nmpc_parameters}. The commanded rotor thrust (converted to rotor speed) obtained from the motion controller is sent to the ESCs via a Kakute H7 Mini autopilot.
	
	\subsection{Benchmark Comparisons}
	\label{sec:benchmark}
	This section presents the benchmark comparisons against our previous work\cite{zhang2022autonomous}. The tests are conducted in real-world environments for more convincing results. We refer readers to the attached video for more details.

\begin{figure}[t]
	\centering
	\includegraphics[width=1\linewidth]{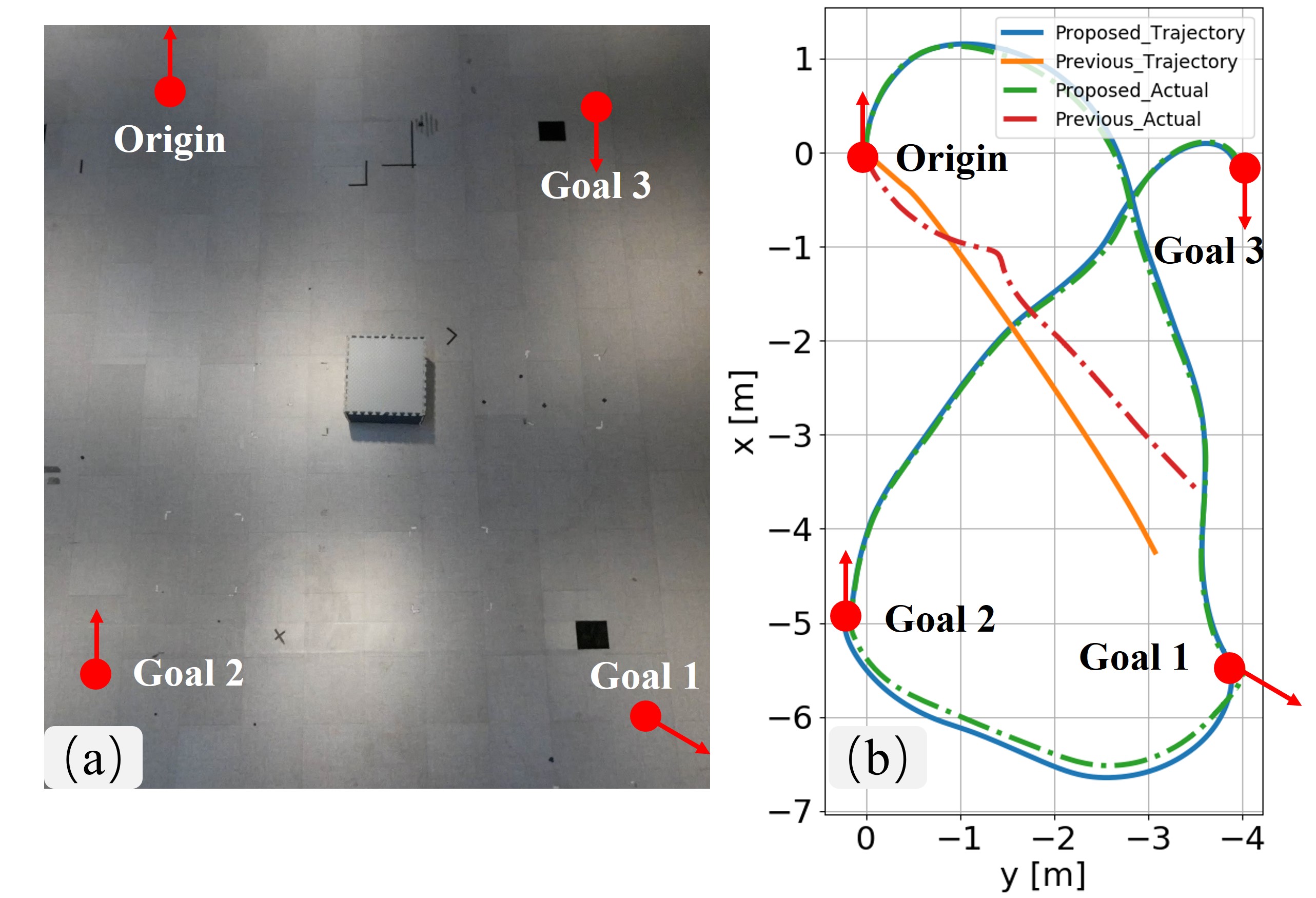}
	\captionsetup{font={small}}
	\caption{
		The trajectory planning benchmark comparison. a) The experimental setting and the given goals. b) The planned and the actual trajectories with the proposed and the previous trajectory planners\cite{zhang2022autonomous}, respectively.
	}
	\label{pic:planning_benchmark_setting}
	\vspace{0cm}
\end{figure}

	\emph{\textbf{1) Comparison of TABV Trajectory Planning:}} Firstly, we compare the proposed TABV trajectory planner with our previous one. As shown in Fig.\ref{pic:planning_benchmark_setting}a, The TABV needs to navigate a series of goals which include the desired heading angles in terrestrial locomotion. These goals are specially arranged so that the TABV cannot proceed in a straight line to the next goal due to the nonholonomic dynamics. This is common in practice, for example, a vehicle needs to turn around when driving toward a rear goal. The planning results of both trajectory planners are tracked by the proposed controller for fair comparison. It turns out that the TABV safely complete the navigation task with the proposed planner, but does not even reach the first goal with our previous planner. As can be seen in Fig.\ref{pic:planning_benchmark_setting}b, the main reason is that the previous planner ignores the nonholonomic dynamics and generates a straight-line trajectory for the sake of smoothness, which is not physically feasible in terrestrial locomotion and eventually leads to the divergence of the controller. In contrast, the proposed planner prevents lateral movement and confines the turning speed within the dynamic limit, thus ensuring the dynamic feasibility.

\begin{table}[]
	\centering
	\caption{TABV Physical Parameters.}
	\setlength{\tabcolsep}{3mm}	
	\begin{tabular}{lc}
		\toprule 
		Parameter  & Value \\ 
		\midrule 
		mass($m$)  [kg] & 0.91 \\ 
		Inertia Matrix($M$)  [gm$^2$] &   diag(7.7, 3.4, 7.3) \\
		arm length($L$)  [m]&  0.23  \\
		rotor thrust coefficient($c_t$) [N] & 1.7$e^{-8}$  \\
		rotor torque coefficient($c_m$)  [Nm$^2$] & 3.7$e^{-10}$ \\
		\bottomrule 
	\end{tabular}
	
	\vspace{-0.2cm}	
	\label{tab:physical_parameters}	
\end{table}  

\begin{table}[]
	\centering
	\caption{NMPC Weights.}
	\setlength{\tabcolsep}{5.5mm}	
	\begin{tabular}{lc}
		\toprule 
		Parameter & Value \\ 
		\midrule 
		$\boldsymbol{W}_{x, p}$ &  diag$\left(8000, 8000, 300\right)$ \\ 
		$\boldsymbol{W}_{x, q}$ &  diag$\left(400, 400, 400, 400\right)$ \\
		$\boldsymbol{W}_{x, v}$ &  diag$\left(100, 100, 100\right)$  \\
		$\boldsymbol{W}_{x, \omega}$ & diag$\left(10, 10, 50\right)$  \\
		$\boldsymbol{W}_{u, T}$ & 0.5 \\
		$\boldsymbol{W}_{u, \tau{xy}}$  & 0.1 \\
		$\boldsymbol{W}_{u, \tau{z}}$  & 0.2 \\
		\bottomrule 
	\end{tabular}
	
	\vspace{-2.7cm}	
	\label{tab:nmpc_parameters}	
\end{table}

	\emph{\textbf{2) Comparison of TABV Tracking Control:}} We then compare the proposed TABV tracking controller with our previous one\cite{zhang2022autonomous}. We choose the root-mean-square-error (RMSE) as the criteria for trajectory tracking performance, which is calculated by
	\begin{equation}
	E_r=\sqrt{\frac{1}{N} \sum_{k=1}^N\left\|\tilde{\boldsymbol{p}}^{\mathrm{k}}-\boldsymbol{p}_{\mathrm{ref}}^{\mathrm{k}}\right\|^2},
	\end{equation}
	where $\tilde{\boldsymbol{p}}^{\mathrm{k}}$ and $\boldsymbol{p}_{\mathrm{ref}}^{\mathrm{k}}$ are the $k$-th sampled actual and desired position, respectively.

\begin{figure}[t]
	\centering
	\includegraphics[width=1\linewidth]{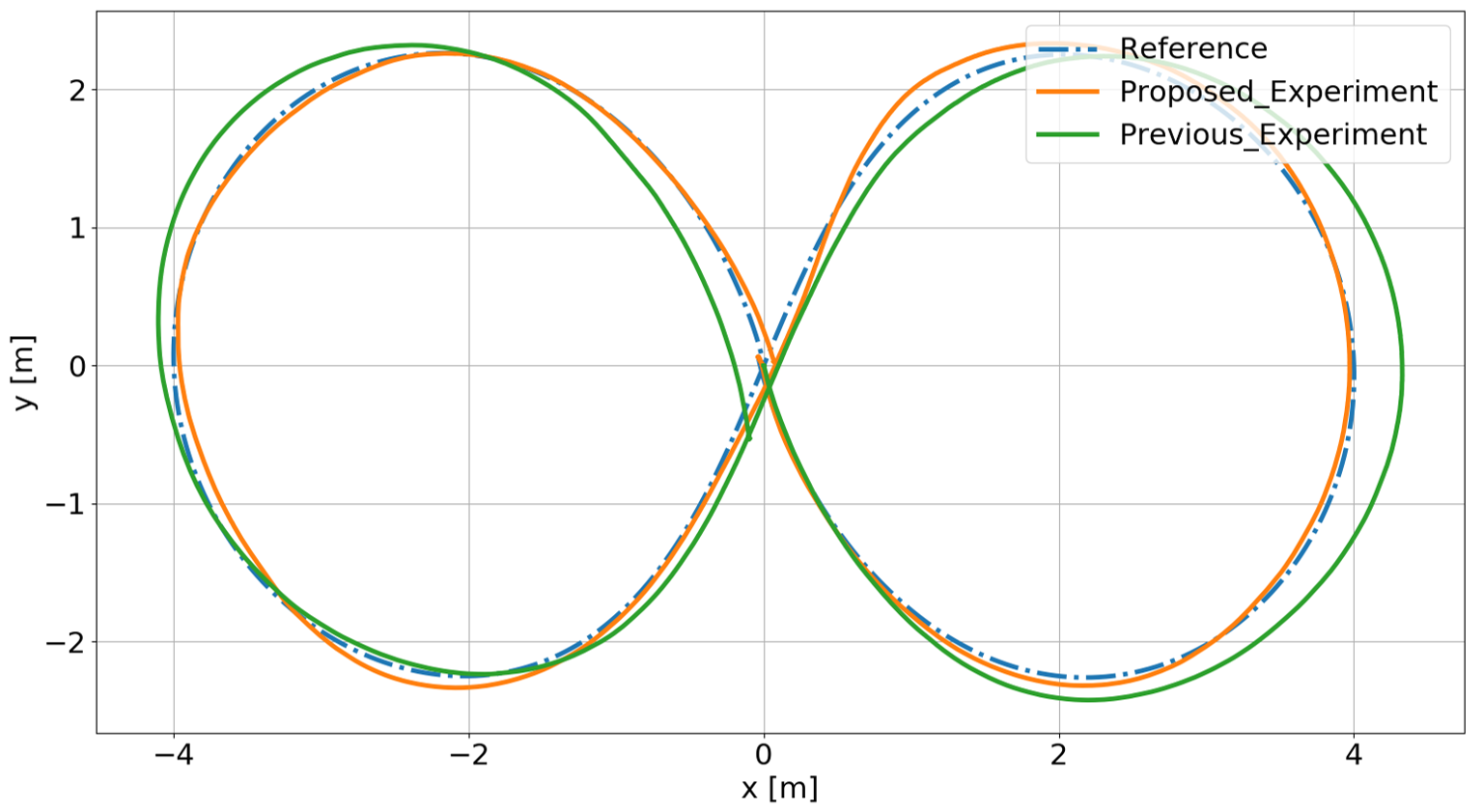}
	\captionsetup{font={small}}
	\caption{
		Comparison of the proposed and the previous controllers\cite{zhang2022autonomous} when tracking a two-dimensional lemniscate trajectory. The RMSE with our proposed and previous controller are $0.11m$ and $0.25m$, respectively.
	}
	\label{pic:control_benchmark_fig}
	\vspace{-0.6cm}
	\end{figure}

	The test is to track a two-dimensional lemniscate trajectory when the  maximal commanded velocity and acceleration limit reach $2.0m/s$ and $1.8m/s^2$, respectively. It can be seen from Fig.\ref{pic:control_benchmark_fig} that when the TABV turns at a high translational speed, the proposed controller still maintains accurate trajectory tracking, while our previous controller does not. The reason is that NMPC considers the TABV's full dynamics, and the predicting nature allows the TABV to react to the turning ahead of time.

	\begin{figure}[t]
	\centering
	\includegraphics[width=1\linewidth]{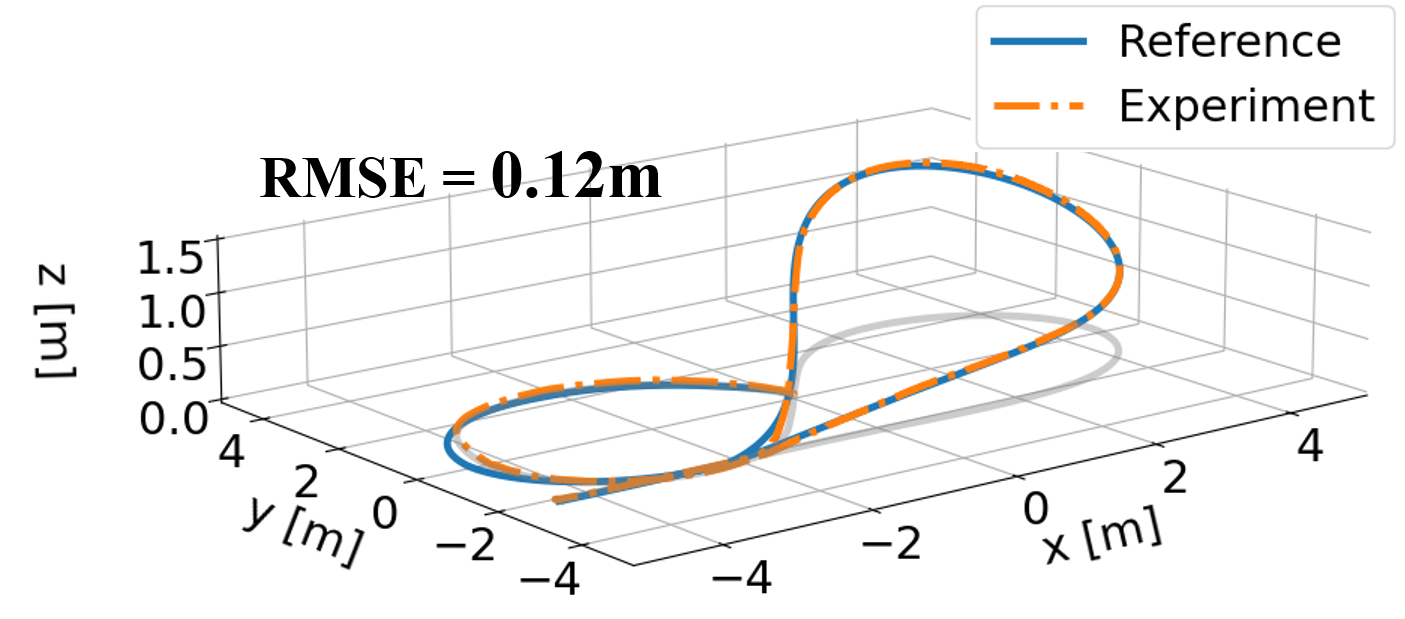}
	\captionsetup{font={small}}
	\caption{
	  The control result when tracking an aggressive terrestrial-aerial hybrid trajectory. The transparent grey curve is the projection of the three-dimensional trajectory on the ground. 
	}
	\label{pic:kongdi_traj}
	\vspace{-0.1cm}
	\end{figure}

	\subsection{Experiments} 
\begin{figure}[t]
	\centering
	\includegraphics[width=1\linewidth]{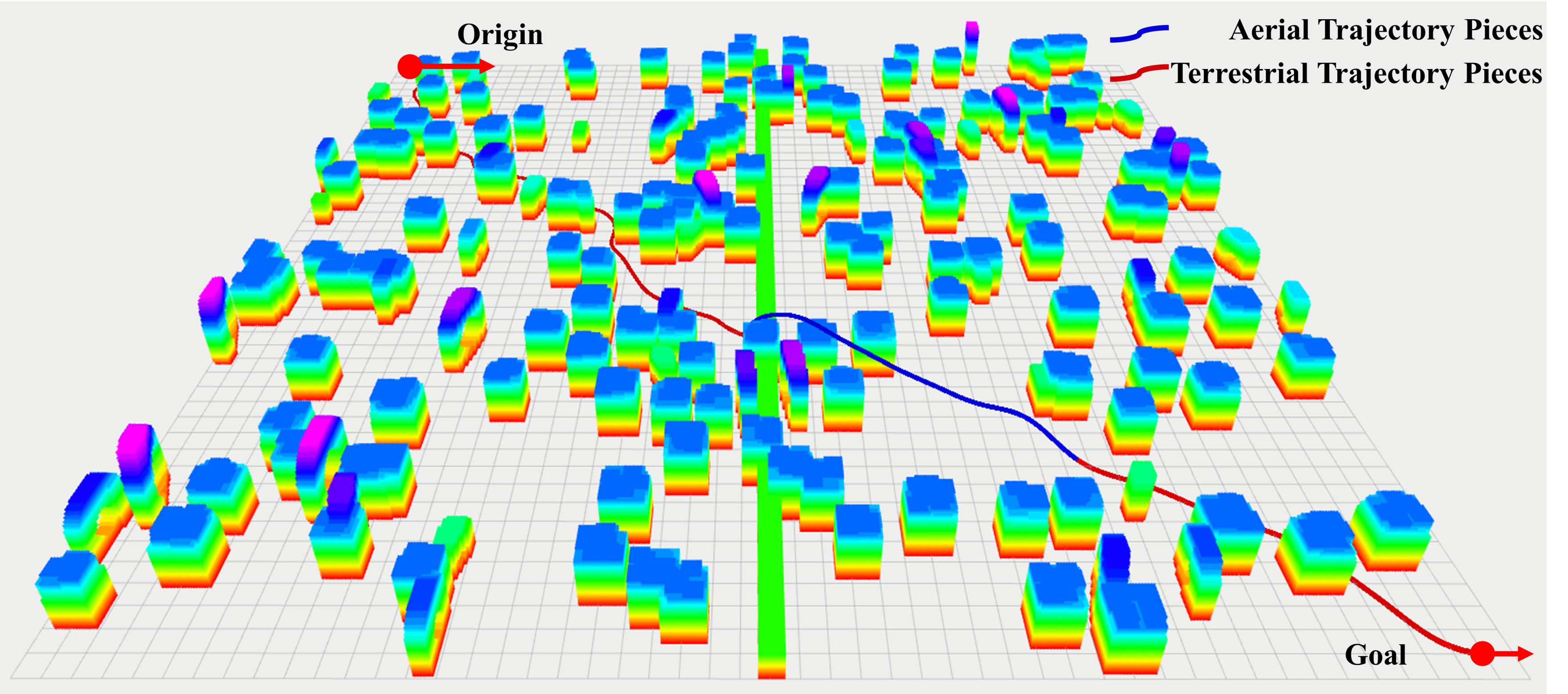}
	\captionsetup{font={small}}
	\caption{
		An instance of the large-scale terrestrial-aerial trajectory planning test.
		The average computation time is $75.4ms$ when generating terrestrial-aerial trajectories with an average length of $76.8m$.
	}
	\label{pic:simu_illustration}
	\vspace{-1.2cm}
\end{figure}

	We also conduct experimental validation of our framework in an unknown complex environment cluttered with obstacles and an insurmountable fence, as shown in Fig.\ref{pic:experiments_scene}a. More details are available in the attached video.
	As a result, the TABV manages to navigate this environment with terrestrial-aerial hybrid locomotion, revealing the robustness of the proposed framework. To further validate the proposed trajectory planner, we test it in a $50m\times50m$ dense simulation environment on an Intel Core i7-10700 CPU. $500$ tests are conducted with randomly arranged obstacles, and an instance of planning results is shown in Fig.\ref{pic:simu_illustration}. As a result, $94\%$ success rate is achieved. The average trajectory length is $76.8m$, while the average  computation times are $14.6ms$ and $70.8ms$, respectively. The above results reveal the proposed methods' robustness and real-time performance.
	
	In addition, to further demonstrate the proposed controller's tracking performance, we conduct a three-dimensional lemniscate trajectory tracking test that requires terrestrial-aerial hybrid locomotion, as illustrated in Fig.\ref{pic:experiments_scene}b. The velocity and acceleration upper bounds reach $3m/s$ and $2.5m/s^2$, respectively. The result shown in Fig.\ref{pic:kongdi_traj} reveals that the proposed controller manages to track the trajectory and achieves smooth mode transition during both takeoff and landing. The RMSE of three-dimensional motion is $0.12m$, which validates the proposed controller's performance when tracking aggressive terrestrial-aerial hybrid trajectories.
	
	\section{Conclusion}
	\label{sec:conclusion}
	In this work, we focus on the autonomous navigation of TABVs and present a model-based planning and control framework. In order to guarantee the physical feasibility and smooth locomotion transition, we first derive a unified TABV dynamic model and its differential flatness to encode the bimodal dynamics in both motion planning and control. Based on the unified dynamic model, a flatness-based trajectory planner and an NMPC-based tracking controller are then proposed. Real-world benchmark comparisons and experiments validate the proposed methods' planning quality and control accuracy.

	In the future, we plan to develop a more comprehensive  planning and control framework that consider terresrial maneuvers on rough terrain. In addition, we will extend the proposed methods to more TABV configurations, so as to make TABVs more widely and effectively used in practice.
	\bibliography{IROS2023_ZRB}
	
\end{document}